\documentclass{ieeeaccess}
\DeclareFontShape{T1}{formata}{m}{sl}{<->ssub * formata/m/it}{}
\usepackage{cite}
\usepackage{amsmath,amssymb,amsfonts}
\usepackage{algorithmic}
\usepackage[pdftex]{graphicx} 
\usepackage{textcomp}

\usepackage{url}
\usepackage{hyperref}
\usepackage{braket}
\usepackage{afterpage}
\usepackage{bm}
\usepackage{mathtools}
\usepackage{multirow}
\usepackage{here}
\usepackage{framed}
\usepackage{tabularx,array}
\usepackage{float}
\newcolumntype{Y}{>{\centering\arraybackslash}X}

\makeatletter
\AtBeginDocument{\DeclareMathVersion{bold}
\SetSymbolFont{operators}{bold}{T1}{times}{b}{n}
\SetSymbolFont{NewLetters}{bold}{T1}{times}{b}{it}
\SetMathAlphabet{\mathrm}{bold}{T1}{times}{b}{n}
\SetMathAlphabet{\mathit}{bold}{T1}{times}{b}{it}
\SetMathAlphabet{\mathbf}{bold}{T1}{times}{b}{n}
\SetMathAlphabet{\mathtt}{bold}{OT1}{pcr}{b}{n}
\SetSymbolFont{symbols}{bold}{OMS}{cmsy}{b}{n}
\renewcommand\boldmath{\@nomath\boldmath\mathversion{bold}}}
\makeatother
\def\BibTeX{{\rm B\kern-.05em{\sc i\kern-.025em b}\kern-.08em
    T\kern-.1667em\lower.7ex\hbox{E}\kern-.125emX}}
\begin{document}
\history{Date of publication xxxx 00, 2026, date of current version xxxx 00, 0000.}
\doi{10.1109/ACCESS.2026.3655591}


\title{SWIFT-FMQA: Enhancing Factorization Machine with Quadratic-Optimization Annealing via Sliding Window}

\author{\uppercase{Mayumi Nakano}\authorrefmark{1},
\uppercase{Yuya Seki}\authorrefmark{1, 2}, \uppercase{Shuta Kikuchi}\authorrefmark{1, 2}, AND \uppercase{Shu Tanaka}\authorrefmark{1, 2, 3, 4}
\IEEEmembership{Member, IEEE}}

\address[1]{Graduate School of Science and Technology, Keio University, Yokohama, Kanagawa 223-8522, Japan}
\address[2]{Keio University Sustainable Quantum Artificial Intelligence Center (KSQAIC), Keio University, Minato-ku, Tokyo 108-8345, Japan}
\address[3]{Department of Applied Physics and Physico-Informatics, Keio University, Yokohama, Kanagawa 223-8522, Japan}
\address[4]{Human Biology-Microbiome-Quantum Research Center (WPI-Bio2Q), Keio University, Shinjuku-ku, Tokyo 160-8582, Japan}
\tfootnote{
This work was partially supported by the Japan Society for the Promotion of Science (JSPS) KAKENHI (Grant Numbers JP23H05447 and 25K07172), the Council for Science, Technology, and Innovation (CSTI) through the Cross-ministerial Strategic Innovation Promotion Program (SIP), ``Promoting the application of advanced quantum technology platforms to social issues'' (Funding agency: QST), Japan Science and Technology Agency (JST) (Grant Number JPMJPF2221). In addition, this paper is partially based on results obtained from a project, JPNP23003, commissioned by the New Energy and Industrial Technology Development Organization (NEDO).
The computations in this work were partially performed using the facilities of the Supercomputer Center, the Institute for Solid State Physics, The University of Tokyo.
M.~N. would like to express his sincere gratitude to The SATOMI Scholarship Foundation for their financial support.
}

\markboth
{M. Nakano \headeretal: SWIFT-FMQA: Enhancing Factorization Machine with Quadratic-Optimization Annealing via Sliding Window}
{M. Nakano \headeretal: SWIFT-FMQA: Enhancing Factorization Machine with Quadratic-Optimization Annealing via Sliding Window}

\corresp{Corresponding author: Mayumi Nakano (e-mail: mayuminakano@keio.jp).}

\newcommand{\STcomment}[1]{{\bf \textcolor{red}{[STcomment:#1]}}}
\newcommand{\STadd}[1]{{\textcolor{red}{#1}}}

\newcommand{\MNcomment}[1]{{\bf \textcolor{blue}{[MNcomment:#1]}}}
\newcommand{\MNadd}[1]{{\textcolor{blue}{#1}}}
\newcommand{\YScomment}[1]{{\bf \textcolor{Turquoise}{[YScomment:#1]}}}

\begin{abstract}
Black-box (BB) optimization problems aim to identify an input that maximizes or minimizes the output of a function (the BB function) whose input-output relationship is unknown.
Factorization machine with quadratic-optimization annealing (FMQA) is a promising approach to this task, employing a factorization machine (FM) as a surrogate model to iteratively guide the solution search via an Ising machine.
Although FMQA has demonstrated strong optimization performance across various applications, its performance often stagnates as the number of optimization iterations increases.
One contributing factor to this stagnation is the growing number of data points in the dataset used to train FM.
As more data are accumulated, the contribution of newly added data points tends to become \textit{diluted} within the entire dataset.
Based on this observation, we hypothesize that such dilution reduces the impact of new data on improving the prediction accuracy of FM.
To address this issue, we propose a novel method named sliding window for iterative factorization training combined with FMQA (SWIFT-FMQA).
This method improves upon FMQA by utilizing a \textit{sliding-window} strategy to sequentially construct a dataset that retains at most a specified number of the most recently added data points.
SWIFT-FMQA is designed to enhance the influence of newly added data points on the surrogate model.
Numerical experiments demonstrate that SWIFT-FMQA obtains lower-cost solutions with fewer BB function evaluations compared to FMQA.
\end{abstract}


\titlepgskip=-21pt

\maketitle

\section{Introduction}
\label{section:Introduction}
\PARstart{C}{ombinatorial} optimization problems seek to find a set of decision variables that satisfy all given constraints and maximize or minimize an objective function. 
Representative examples of combinatorial optimization problems include the traveling salesman problem, the quadratic assignment problem, the quadratic knapsack problem, and the low autocorrelation binary sequences (LABS) problem.
The problem size of combinatorial optimization problems is characterized by the number of decision variables.
As the number of decision variables increases, the number of candidate solutions grows exponentially, leading to a combinatorial explosion, which makes exhaustive search for optimal solutions difficult.
In this context, research on metaheuristics that can find high-quality solutions in a short time has progressed.
Recently, Ising machines, which are hardware implementations of algorithms such as simulated annealing~\cite{Kirkpatrick1983Optimization, Johnson1989Optimization, Johnson1991Optimization} and quantum annealing~\cite{Kadowaki1998Quantum, Das2008Colloquium, Tanaka2017Quantum, Chakrabarti2023Quantum}, both being types of metaheuristics, have attracted significant attention~\cite{Mohseni2022Ising, Yulianti2022Implementation, Jiang2023Classifying, Kikuchi2025Effectiveness}.
Ising machines are capable of sampling good solutions in a wide range of applications, including financial portfolio optimization~\cite{Phillipson2021Portfolio, Sakuler2025Real}, online advertisement~\cite{Tanahashi2019Application}, traffic~\cite{Neukart2017Traffic, Stollenwerk2019Quantum, Inoue2021Traffic, Mukasa2021Ising, Marchesin2023Improving, Kanai2024Annealing}, materials simulation~\cite{Harris2018Phase, King2018Observation, Utimula2021Quantum, Sampei2023Quantum}, and computer-aided engineering~\cite{Endo2022Phase, Honda2024Development, Xu2025Quantum}.
When solving combinatorial optimization problems using an Ising machine, it is necessary to formulate the problem as an energy function of an Ising model or in the form of quadratic unconstrained binary optimization (QUBO).
Various combinatorial optimization problems can be represented by these formulations~\cite{Lucas2014Ising}.

Among combinatorial optimization problems, some are difficult to formulate as an Ising model or QUBO.
A typical example is the black-box (BB) optimization problem.
A characteristic of BB optimization problems is that the objective function is not explicitly given.
This situation arises when obtaining the output value of the objective function requires conducting simulations or experiments.
For example, applications include drug design with affinity for specific protein targets~\cite{Lamanna2023GENERA}, optimization of interfacial thermal conductivity in nanostructured materials~\cite{Ju2017Designing}, and the design of convolutional neural network cell architectures~\cite{Habi2019Genetic}.
The objective function in BB optimization problems is called a black-box (BB) function.
In real-world problems such as those described above, calling the BB function incurs time and financial costs.
Therefore, it is required to conduct BB optimization while minimizing the number of BB function evaluations.
Under the drawback that the objective function is not explicitly defined, Ising machines cannot solve these problems.

Factorization machine with quantum annealing (FMQA) has been proposed to overcome this drawback~\cite{Kitai2020Designing}.
FMQA is a BB optimization method that iteratively performs four steps: prediction of the BB function using a machine learning model called factorization machine (FM)~\cite{Rendle2010FactorizationM} as a surrogate model, sampling solutions using an Ising machine, calling the BB function for evaluation, and constructing the dataset for FM training.
FMQA is collectively referred to as factorization machine with quadratic-optimization annealing when the employed Ising machine does not utilize quantum technologies~\cite{Tamura2025Black}.
FMQA solves the problem of implicitly given objective functions by approximately formulating the BB function in the QUBO form using FM.
Additionally, by utilizing FM as a surrogate model instead of directly using the BB function, FMQA can reduce the number of BB function evaluations.
FMQA has been reported to be capable of searching for lower-cost solutions than conventional BB optimization methods.
For instance, in the structural design of photonic crystal lasers~\cite{Inoue2022Towards}, FMQA outperforms the genetic algorithms and particle swarm optimization method.
In addition, FMQA has proven superior to Bayesian optimization in barrier material design~\cite{Nawa2023Quantum} and printed circuit board design~\cite{Matsumori2022Application}.
FMQA iteratively generates data points by using an Ising machine for FM training, and they are added to the initial dataset.
This increase in training data is expected to improve FM's prediction accuracy, thereby leading to better solutions.

In contrast, it has been reported that FMQA tends to experience stagnation in solution improvement as the optimization process iterates~\cite{Ross2023Hybrid, Maruo2025Topology}.
To solve this issue, we focused on the dynamics of FM in the FMQA process.
It is necessary to improve the prediction accuracy of FM for low-cost solutions of the BB function for better optimization performance of FMQA.
When the solution is improved by FMQA, the parameters of the FM are updated by adding training data, which improves the prediction accuracy.
However, depending on the training data, the prediction accuracy for low-cost solutions does not improve, and the improvement of the solution stagnates.
One reason for this issue is the excessive increase in the number of data points in the dataset used for FM training as the optimization process iterates.
Even when solutions that could improve the prediction accuracy of FM are added to the dataset, the contribution of the newly added solutions is ignored because of the large number of existing data points; that is, the new solutions are \textit{diluted}.
We infer that this dilution of the solutions is one of the causes of the stagnation in solution improvement.

\begin{figure*}[t]
    \centering
    \includegraphics[width=1.0\textwidth]{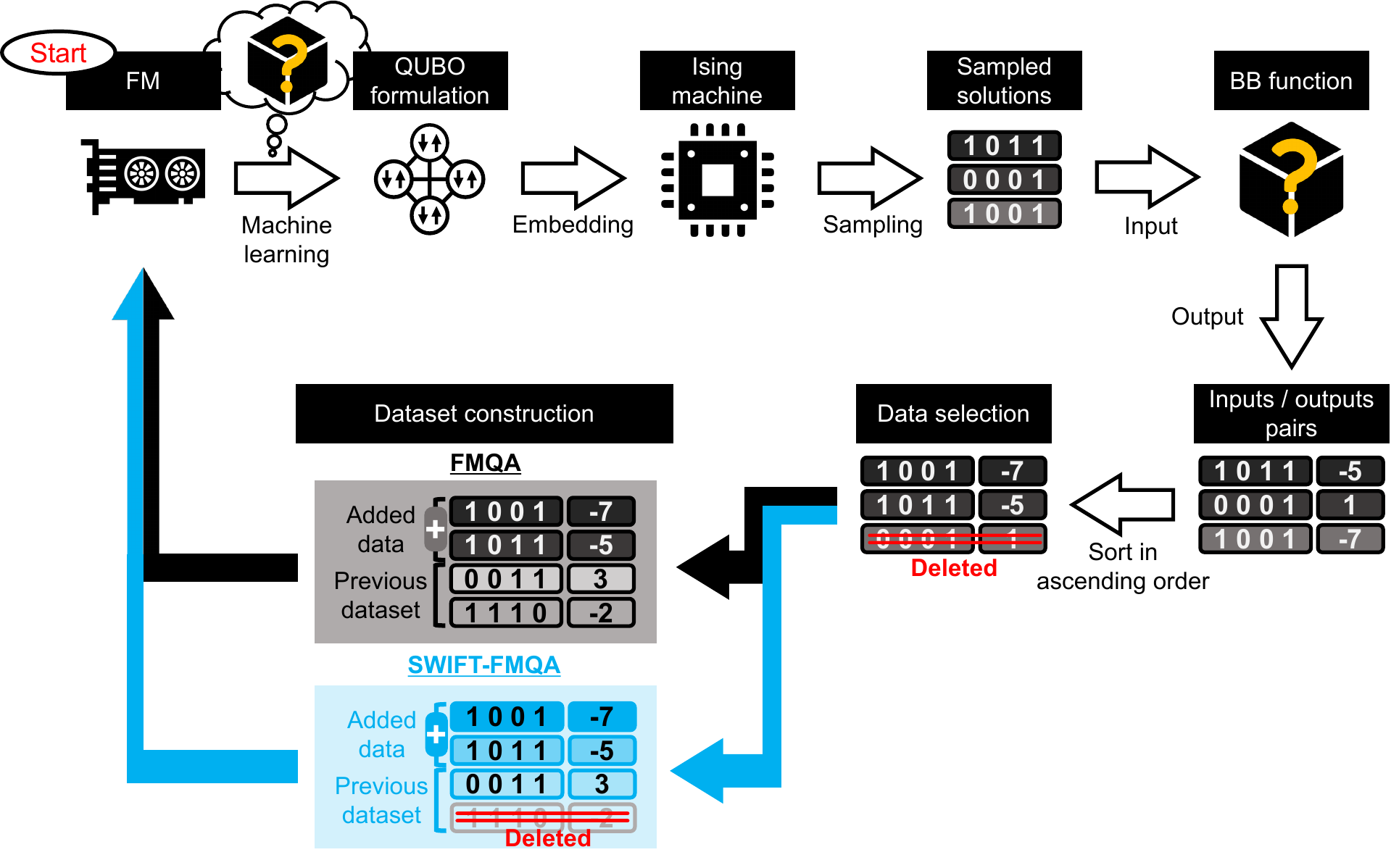}
    \caption{Illustration of the flow of FMQA, including SWIFT-FMQA. Unlike FMQA, SWIFT-FMQA sets an upper limit on the number of data points in the dataset; therefore, some existing data points are discarded during the dataset construction.}
    \label{illust:FMQA}
\end{figure*}

In this study, we propose a novel method named sliding window for iterative factorization training combined with FMQA (SWIFT-FMQA) (Fig.~\ref{illust:FMQA}).
SWIFT-FMQA enhances FMQA by employing a \textit{sliding-window} strategy~\cite{Widmer1996Learning, Iwashita2018Overview} to sequentially construct a dataset that retains at most a specified number of the most recently added data points.
By setting the upper limit on the number of data points, we can prevent the dilution of solutions.
This study aims to numerically demonstrate that SWIFT-FMQA prevents stagnation in the improvement of solutions and enhances the optimization performance of FMQA.
To this end, we performed BB optimization using SWIFT-FMQA, treating the LABS problem, known as a benchmark problem in combinatorial optimization, as a BB optimization problem.
As a result, we show that SWIFT-FMQA searched for lower-cost solutions with fewer BB function evaluations compared to FMQA.

This paper consists of the following sections.
Section~\ref{section:FMQA} describes the overview of FMQA.
Section~\ref{section:SWIFT-FMQA} explains the motivation for introducing SWIFT-FMQA and describes its procedure.
Section~\ref{section:Simulation settings} describes the setup for the numerical experiments.
Section~\ref{section:Results} presents the results of the optimization performance of SWIFT-FMQA.
Here, we show that SWIFT-FMQA exhibits high optimization performance for various values of the hyperparameter of FM and the number of data points in the initial dataset.
Section~\ref{section:Discussion} is dedicated to discussing the optimization performance of SWIFT-FMQA.
Section~\ref{section:Conclusion} presents the conclusion.

\section{FMQA}
\label{section:FMQA}
In this section, we describe the flow of FMQA.
FMQA consists of four main steps: prediction of the BB function using FM, a machine learning model, as a surrogate model; sampling of solutions using an Ising machine; evaluation of the solutions by calling the BB function; and the construction of the dataset for FM training.
The procedure described here is fundamentally based on the framework proposed by Kitai et al.~\cite{Kitai2020Designing}.
However, it differs in the data selection strategy employed in the fourth step for the dataset construction.
Whereas the original proposal~\cite{Kitai2020Designing} and subsequent variants~\cite{Nawa2023Quantum, Inoue2022Towards} typically select data based on values predicted by FM, we instead adopt a strategy that selects data according to BB function values, following the approach introduced in~\cite{Couzinié2025Machine}.
By evaluating data with the BB function before adding them to the dataset, we prioritize the reliability of the training data. 
The details of the steps are described below.

Only at the beginning of the optimization process, a dataset for the initial training of FM needs to be prepared manually.
The details of all the steps, including the preparation of the initial training dataset, are described below.

\subsection{Preparation: constructing the dataset for initial training of FM}
\label{subsection:Preparation: constructing the dataset for initial training of FM}
To construct the dataset for the initial training, $D_{\mathrm{init}}$ inputs to the BB function $(\bm{x}^{(1)}, \dotsc, \bm{x}^{(D_{\mathrm{init}})})$ are prepared.
Here, each input is an $N$-dimensional binary sequence: $\bm{x}^{(d)}\in \{0, 1\}^N$ $(d =1, \dotsc, D_{\mathrm{init}})$.
Then, the output of the BB function $y^{(d)}=f(\bm{x}^{(d)})$ is obtained.
The dataset for initial training is denoted as 
\begin{equation}
\label{eq:initial_dataset}
\mathcal{D} \equiv \{(\bm{x}^{(1)}, y^{(1)}), \dotsc, (\bm{x}^{(D_{\mathrm{init}})}, y^{(D_{\mathrm{init}})})\}.
\end{equation}
This dataset $\mathcal{D}$ is used in Step 1 for the prediction of the BB function by FM.
Note that this preparation step is performed only once in this optimization process.

\subsection{Step 1: prediction of the BB function by FM}
\label{Step 1: prediction of the BB function by FM}
FM is trained with the dataset $\mathcal{D}$.
The model equation of FM is defined as 
\begin{equation}
\label{eq:FM-general}
f_{\mathrm{FM}}(\boldsymbol{x}; \bm{\theta}) =\omega_{0}+\sum_{i=1}^{N}\omega_{i}x_{i}+\sum_{1\le i < j \le N}\langle \bm{v}_i, \bm{v}_j \rangle x_{i}x_{j},
\end{equation}
where $\bm{x} = (x_1, \dotsc , x_N) \in \{0, 1\}^N$, and $\omega_0 \in \mathbb{R}$, $\bm \omega = (\omega_1, \dotsc, \omega_N)  \in \mathbb{R}^N$, and $\bm{v}_i \in \mathbb{R}^K$ are the model parameters.
The parameter $K \in \mathbb{N}$ is the hyperparameter of FM.
All the model parameters are denoted by $\bm{\theta}=(\omega_0, \bm \omega, \{\bm v_i \})$ for simplicity.
Here, $\langle \cdot, \cdot \rangle$ represents the inner product.
In this step, the model parameters are updated to minimize the loss function.
In this study, we used the mean squared error (MSE) as the loss function:
\begin{equation}
\label{eq:learning_loss}
L(\bm \theta) \equiv \frac{1}{D} \sum_{d=1}^{D} \left[{{f_{\mathrm{FM}}}(\bm {x}^{(d)};{\bm \theta})-f({\bm {x}^{(d)}})}\right]^2,
\end{equation}
where $D$ represents the number of data points in the dataset $\mathcal{D}$.

Regarding the hyperparameter $K$, a small value $K < N$ is commonly adopted.
Setting a large $K$ increases the number of model parameters and improves the model's prediction accuracy. 
However, it has been pointed out that if $K$ is too large, overfitting may occur.
Appropriate selection of $K$ is required~\cite{Rendle2010FactorizationM}.

\subsection{Step 2: sampling solutions using an Ising machine}
\label{Step 2: sampling solutions using an Ising machine}
In this step, low-cost solutions of the predicted FM are sampled using an Ising machine.
The model equation of FM in~\eqref{eq:FM-general} is in the QUBO form.
Here, QUBO is a quadratic expression of binary variables as shown below:
\begin{equation}
\label{eq:QUBO-general}
\mathcal{H} = c + \sum_{i=1}^N Q_{i, i}x_i + \sum_{1\le i < j \le N}Q_{i,j}x_{i}x_{j}.
\end{equation}
Here, $c \in \mathbb{R}$ is a constant, $\bm{x}=(x_1, \dots, x_N)\in \{0, 1\}^N$, and $N$ is the number of binary variables $\bm{x}$.
An upper triangular matrix $Q \in \mathbb{R}^{N \times N}$ whose $(i, j)$ component is $Q_{i, j}$ is called the QUBO matrix.
Comparing~\eqref{eq:FM-general} with~\eqref{eq:QUBO-general}, it can be seen that the model equation of FM is in the QUBO form.
Hence, low-cost solutions to FM can be sampled by an Ising machine.

The solutions sampled at this step are expected to minimize the BB function.
This is because the FM constructed in Step 1 is expected to approximate the BB function accurately.
We sample $D_{\mathrm{reads}}$ low-cost solutions to the FM $\bm{x}^* = \{\bm{x}^{(1)}, \dotsc, \bm{x}^{(D_{\mathrm{reads}})}\}$ using an Ising machine.

\subsection{Step 3: calling the BB function for evaluation}
\label{Step 3: calling BB function for evaluation}
In this step, we evaluate the solutions obtained in Step 2 by calling the BB function.
The BB function is called for each solution in $\bm{x}^*$, obtaining the corresponding outputs.
Accordingly, we obtain $D_{\mathrm{reads}}$ input-output pairs $(\bm{x}^*, \bm{y}^*)=\{(\bm{x}^{(1)}, y^{(1)}), \dotsc, (\bm{x}^{(D_{\mathrm{reads}})}, y^{(D_{\mathrm{reads}})})\}$.

\subsection{Step 4: constructing the dataset for FM training}
\label{Step 4: constructing the dataset for FM training}
In this step, the dataset $\mathcal{D}$ is updated by adding the input-output pairs obtained in Step 3.
Before adding to the dataset, we extract a subset of $D_{\mathrm{adds}}$ pairs from the $D_{\mathrm{reads}}$ obtained, where $D_{\mathrm{adds}} \leq D_{\mathrm{reads}}$.
We first sort $\bm{y}^*$ in ascending order and select the $D_{\mathrm{adds}}$ lowest values and their corresponding inputs.
Thus, the selected set is given by:
\begin{equation}
\label{eq:selected_Dadds}
\begin{split}
(\bm{x}^*, \bm{y}^*) &= \{(\bm{x}^{(1)}, y^{(1)}), \dotsc , (\bm{x}^{(D_{\mathrm{adds}})}, y^{(D_{\mathrm{adds}})})\}.
\end{split}
\end{equation}
Then, the dataset $\mathcal{D}$ is updated by adding the input-output pairs $(\bm{x}^*, \bm{y}^*)$.
This sorting strategy has been adopted in a previous study~\cite{Couzinié2025Machine}.
By preferentially adding low-cost solutions to the dataset, we aim to improve the accuracy of prediction by FM for the neighborhood of low-cost regions within the BB function's energy landscape.
In FMQA, the input-output pairs are simply added to the dataset.
Then, the process returns to Step 1 and Steps 1--4 are iterated for a preset number $N_{\mathrm{iter}}$, causing the increase of the number of data points in the dataset.

Through the iterative execution of these steps, the prediction accuracy of FM is expected to improve, which in turn facilitates the sampling of low-cost solutions by an Ising machine.

\section{SWIFT-FMQA}
\label{section:SWIFT-FMQA}
In this section, we first elucidate the theoretical motivation behind SWIFT-FMQA, focusing on the issue of data dilution in FMQA.
Subsequently, we describe the procedure of SWIFT-FMQA.

\subsection{Motivation: data dilution in FMQA}
\label{subsection: Motivation: data dilution in FMQA}
As described in Section \ref{section:FMQA}, the model parameters of FM are updated to minimize the loss function $L(\bm \theta)$ shown in~\eqref{eq:learning_loss} for $D$ data points in the prepared dataset $\mathcal{D}$.
Consider the case where a single data point is added to the dataset $\mathcal{D}$.
The loss function $L(\bm \theta)$ is expressed as follows using~\eqref{eq:learning_loss}.
\begin{equation}
\begin{split}
\label{eq:learning_loss_revised}
L(\bm{\theta}) &= \frac{1}{D+1} \Biggl[ \sum_{d=1}^{D} \left( f_{\mathrm{FM}}(\bm{x}^{(d)}; \bm{\theta}) - f(\bm{x}^{(d)}) \right)^2 \\
&\hphantom{=} + \left( f_{\mathrm{FM}}(\bm{x}^{(D+1)}; \bm{\theta}) - f(\bm{x}^{(D+1)}) \right)^2 \Biggr].
\end{split}
\end{equation}
Here, the second term on the right-hand side represents the loss function for a data point sampled by an Ising machine in a single optimization iteration.
When the number of data points $D$ in the dataset becomes large, such as in the later stages of optimization iteration, the contribution of a newly added solution to the loss function $L(\bm{\theta})$ is diluted by the $1/(D+1)$ factor.
As a result, the gradient used to update the model parameters is dominated by the large amount of past data.
This leads to a form of high model \textit{inertia}, where the model becomes resistant to changes even when a new, potentially more informative solution is added to the dataset.
We hypothesize that this dilution is a cause of the optimization stagnation observed in FMQA.

To mitigate this dilution effect, it is essential to impose an upper limit on the number of data points $D$, thereby ensuring that the influence of the newly sampled solutions remains significant.
Furthermore, as solutions obtained in the later stages of the optimization iterations are generally more promising, prioritizing the retention of recently added solutions while discarding older ones is effective.
These considerations motivate the adoption of the sliding-window strategy implemented in SWIFT-FMQA.
The sliding-window strategy is a common approach for addressing concept drift, where data characteristics change dynamically over time.
While sliding window can employ either a fixed window size~\cite{Widmer1996Learning, Iwashita2018Overview} or an adaptive window size~\cite{Klinkenberg1998Adaptive}, we adopt the former.
The fixed sliding-window approach is widely adopted in machine learning studies dealing with sequential data, such as calendar scheduling~\cite{Mitchell1994Experience}, credit card fraud detection~\cite{Jurgovsky2018Sequence}, and protein feature extraction~\cite{Zhang2025BioSemAF}. 
Furthermore, a method combining a sliding window with FMQA has been proposed for dynamic discrete environments and real-time BB optimization~\cite{Kashimata2025Real}.
However, a key distinction lies in the objective: while the authors in~\cite{Kashimata2025Real} employ the sliding window to adapt to dynamically changing environments, SWIFT-FMQA utilizes it to mitigate data dilution.

\subsection{Procedure of SWIFT-FMQA}
\label{subsection: Procedure of SWIFT-FMQA}
\begin{figure*}[t]
    \centering
    \includegraphics[width=0.8\textwidth]{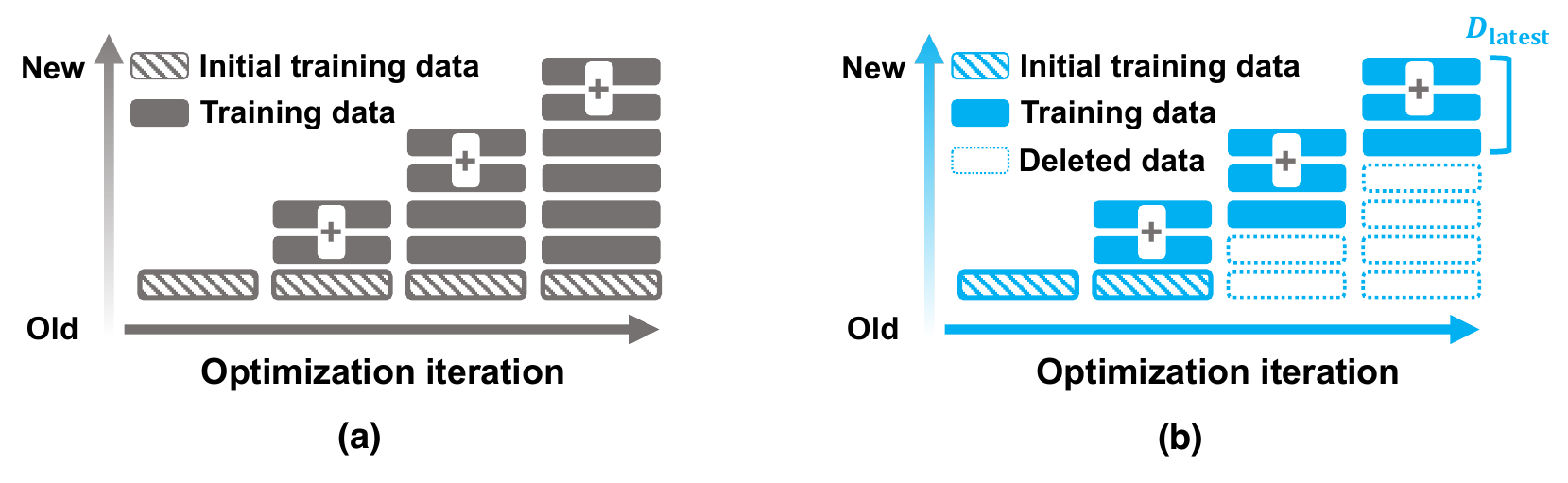}
    \caption{Comparison in terms of the dataset construction between (a) FMQA and (b) SWIFT-FMQA. The plus marks represent $D_{\mathrm{adds}}$ solutions added to the dataset in one optimization iteration of FMQA and SWIFT-FMQA. The total number of optimization iterations is $N_{\mathrm{iter}}$. $D_{\mathrm{latest}}$ represents the upper bound on the number of data points in the dataset in SWIFT-FMQA. In this illustration, $D_{\mathrm{adds}}=2$ and $D_{\mathrm{latest}}=3$.}
    \label{illust:Comparison_FMQA}
\end{figure*}
SWIFT-FMQA serves as an improvement to Step 4 of FMQA: constructing the dataset for FM training.
To formulate SWIFT-FMQA, let $t = 1, \dotsc ,N_{\mathrm{iter}}$ denote the index of the current optimization iteration, and $\mathcal{D}^{(t)}$ be the dataset obtained after update in Step 4 of the $t$-th iteration.
Consequently, the dataset used for FM training in Step 1 of the $t$-th iteration corresponds to $\mathcal{D}^{(t-1)}$.
Note that $\mathcal{D}^{(0)}$ is defined as the dataset for initial training in \eqref{eq:initial_dataset}.
To provide a generalized formulation that accommodates cases where duplicate data points are allowed, we define the dataset used in this study as a multiset.
The potential impact of allowing duplicate data points is discussed in Section~\ref{section:Discussion}.
In addition, let $D^{(t)}$ be the number of data points in $\mathcal{D}^{(t)}$, and $\mathcal{S}^{(t)}$ be the set of $D_{\mathrm{adds}}$ data points to be added to the dataset $\mathcal{D}^{(t-1)}$.
The set $\mathcal{S}^{(t)}$ corresponds to the selected set defined in \eqref{eq:selected_Dadds}.

Fig.~\ref{illust:Comparison_FMQA} illustrates a comparison of FMQA and SWIFT-FMQA for the dataset construction.
In FMQA, the dataset is updated simply by taking the multiset sum of the current dataset and the selected set:
\begin{equation}
\label{eq:FMQA_dataset_update}
\mathcal{D}^{(t)} = \mathcal{D}^{(t-1)} + \mathcal{S}^{(t)}.
\end{equation}
Fig.~\ref{illust:Comparison_FMQA}(a) shows that the number of data points in the dataset $D$ increases as the optimization iterations proceed in FMQA.
In contrast, SWIFT-FMQA restricts the number of data points in the dataset using a sliding-window strategy with a window size of $D_{\mathrm{latest}}\in \mathbb{N}$, as illustrated in Fig.~\ref{illust:Comparison_FMQA}(b), whereby data points outside the window are discarded.
In the first optimization iteration, two solutions are added to the dataset, and since all data points lie within the window, none are discarded.
In the subsequent second and third iterations, two solutions are added at each step, while two existing data points are simultaneously discarded.
In this way, SWIFT-FMQA maintains a dataset containing at most $D_{\mathrm{latest}}$ data points. 

The update rule for the dataset in SWIFT-FMQA is formulated as:
\begin{equation}
\label{eq:SWIFT-FMQA_dataset_update}
\mathcal{D}^{(t)} = \mathcal{D}^{(t-1)} + \mathcal{S}^{(t)} - \mathcal{D}_{\mathrm{old}}^{(t)},
\end{equation}
where the minus sign denotes the multiset difference operator.
The multiset $\mathcal{D}_{\mathrm{old}}^{(t)}$ contains the oldest data points in the multiset $\mathcal{D}^{(t-1)} + \mathcal{S}^{(t)}$.
The number of data points in $\mathcal{D}_{\mathrm{old}}^{(t)}$ (i.e., the number of data points to be discarded) is determined as:
\begin{equation}
\label{eq:propFMQA_D_old}
D_{\mathrm{old}}^{(t)} =
\max \{D^{(t-1)} + D_{\mathrm{adds}} - D_{\mathrm{latest}}, 0\}.
\end{equation}

\section{Simulation settings}
\label{section:Simulation settings}
In this section, we describe the settings of the BB function used in the numerical experiments, along with the configurations of FM and the Ising machine, to evaluate the optimization performance of SWIFT-FMQA.

\subsection{Black-box function}
\label{subsection: Black-box function}
We adopted the LABS problem as the BB optimization benchmark.
The LABS problem~\cite{Schroeder1970Synthesis, Golay1977Sieves, Hoholdt1988Determination, Bernasconi1987Low} is a combinatorial optimization problem inherent in various real-world applications.
For instance, it is applied in side lobe suppression in radar and sonar~\cite{Beenker1985Binary}, enhancing the security of stream cipher key sequences~\cite{Cai2009Binary}, and improving the measurement accuracy of impulse responses in oil well exploration~\cite{Song2018Analysis}.

The objective of the LABS problem is to find the sequences that minimize the sum of squares of the autocorrelation function of a sequence consisting of $N$ spin variables $s_i\in \{-1, +1\}$ $(i=1, \dotsc , N)$.
The autocorrelation function is defined as
\begin{equation}
\label{eq:labs_autocorrelation}
C_k(\bm s) = \sum_{i=1}^{N-k} s_i s_{i+k}, \quad \bm{s} = (s_1, s_2, ..., s_N).
\end{equation}
Furthermore,~\eqref{eq:labs_autocorrelation} leads the sum of squares of the autocorrelation function $E(\bm{s})$ to be defined as
\begin{equation}
\label{eq:labs_energy}
E(\bm{s}) = \sum_{k=1}^{N-1}C_k^2(\bm{s}).
\end{equation}
Using~\eqref{eq:labs_energy}, the merit factor $F(\bm{s})$~\cite{Golay1977Sieves, Golay1982merit, Golay1990new} is introduced as
\begin{equation}
\label{eq:labs_merit_factor}
F(\bm{s}) = \frac{N^2}{2E(\bm{s})}.
\end{equation}
Therefore, the LABS problem is defined as
\begin{align}
\label{eq:labs_formulation}
\mathrm{minimize} \ {E(\bm{s})} \
\nonumber
\mathrm{or} \
\mathrm{maximize} \ {F(\bm{s})}.
\end{align}

In this study, the maximization problem of the merit factor $F(\bm{s})$ was treated as a BB optimization problem.
Here, in BB optimization with FMQA and SWIFT-FMQA, the maximization problem is reformulated as a minimization problem by treating $-F(\bm{s})$ as the BB function.
Although the BB function is defined over spin variables $\bm{s}$, the solutions sampled in Step 3 of the FMQA optimization process (Section~\ref{Step 3: calling BB function for evaluation}) are expressed in terms of binary variables $\bm{x}$.
Accordingly, the binary variables must be converted into spin variables using the following transformation:
\begin{equation}
\label{eq:binary_to_spin}
s_i = 2x_{i}-1.
\end{equation}

The first reason that we selected the LABS problem as a BB optimization problem is that conducting BB optimization of the LABS problem using FMQA is a challenging task.
This challenging task can be specifically divided into issues related to the form of the LABS problem's equation and its energy landscape.
Firstly, in terms of the form of the LABS problem's equation, the number of candidate solutions for the LABS problem is $2^N$, and the computational complexity for an exhaustive search is $\mathcal{O}(2^N)$.
Thus, finding the exact solution through exhaustive search becomes exponentially difficult as $N$ increases.
Additionally, the negative merit factor $-F(\bm{s})$ is composed of the reciprocal of a quartic polynomial in the spin variables.
As a result, predicting the BB function with FM, which can model interactions only up to quadratic order, becomes fundamentally difficult.
This structural mismatch means that the surrogate model cannot theoretically capture the exact landscape of the BB function, regardless of the amount of training data.
This situation closely resembles real-world BB optimization problems, where the underlying physics or logic is often far more complex than what a simple quadratic surrogate model can represent.
Therefore, demonstrating high optimization performance on the LABS problem serves as strong evidence of the robustness of FMQA or SWIFT-FMQA against model misspecification and its capability to extract essential features from a complex landscape.
Secondly, regarding the energy landscape of the LABS problem, it exhibits a ``bit-flip'' neighborhood structure where a single bit flip can cause the objective function value to fluctuate by tens of percent~\cite{Kratica2012Electromagnetism}.
As also stated in Section~\ref{section:Introduction}, FMQA has been reported to exhibit stagnation in solution improvement and get trapped in local optima as the optimization process iterates~\cite{Ross2023Hybrid, Maruo2025Topology}.
By treating the LABS problem as a BB optimization problem, we believe that the effectiveness of SWIFT-FMQA against the challenges of FMQA can be confirmed.

The second reason is the ease of quantitative evaluation of the sampled solutions.
In the previous research, optimal solutions by exhaustive search for $N \leq 66$ and the skewsymmetric optimal solutions for $N \leq 119$ of the LABS problem have been reported~\cite{Packebusch2016Low}.
These known results enable a precise assessment of the objective function values of the solutions generated by SWIFT-FMQA.
For these two reasons, the LABS problem has been widely adopted as a standard benchmark for various optimization methods~\cite{Gent1999CSPLIB}, including quantum optimization~\cite{Koch2025Quantum}, and is therefore considered an appropriate choice as the BB optimization problem in this study.

\subsection{Settings for FMQA and SWIFT-FMQA}
\label{subsection: FMQA / SWIFT-FMQA settings}

Table~\ref{table:Parameter_for_FMQA} presents the FMQA and SWIFT-FMQA parameter settings.
Regarding the FM settings, the adaptive moment estimation with weight decay (AdamW) algorithm~\cite{Loshchilov2017Decoupled} was adopted for the optimizer.
We used the default parameters for AdamW proposed in ~\cite{Loshchilov2017Decoupled}.

For the initial training data, $D_{\mathrm{init}}$ unique binary arrays were randomly generated from a uniform distribution.
In SWIFT-FMQA, these $D_{\mathrm{init}}$ data points were used for the first round of FM training, regardless of the specified number of data points $D_{\mathrm{latest}}$ used in later training.
That is, even when $D_{\mathrm{latest}} < D_{\mathrm{init}}$, the initial training was conducted using all $D_{\mathrm{init}}$ data points, and the use of $D_{\mathrm{latest}}$ data points began from the second training iteration onward.

As mentioned in Section~\ref{section:FMQA}, the model parameters $\bm{\theta}$ of FM shown in~\eqref{eq:FM-general} are updated each time FM is trained.
Note that, the initial values of these parameters are set manually only during initial training.
In this study, we initialized the model parameters so that the expected value of the initial FM in~\eqref{eq:FM-general} is zero.
Specifically, the initial value for $\omega_0$ was set to zero.
The model parameters $\bm\omega$ and $\{\bm{v}_i\}$ are sampled from a uniform distribution with individual ranges $[-L_1, L_1)$ and $[-L_2, L_2)$, respectively.
Here, $L_1$ ($L_2$) was set so that the variance of the linear (quadratic) term in~\eqref{eq:FM-general} approaches unity.

For the Ising machine, we used a software-based Ising solver called Simulated Annealing Sampler~\cite{D-Wave2022dwave}, developed by D-Wave Systems.
Hereafter, the term Ising solver refers specifically to Ising machines implemented in software rather than hardware.
The settings for the initial inverse temperature, final inverse temperature, outer loop count, and inner loop count in the Ising solver were determined greedily, adopting what was considered the optimal configuration.
\begin{table}[t]
\caption{Parameter settings for FMQA and SWIFT-FMQA.}
\label{table:Parameter_for_FMQA}
\centering
    \begin{tabular}{|l||c|}
        \hline
        \multicolumn{2}{|c|}{\textbf{Machine learning model: FM}} \\ \hline \hline
        Hyperparameter $K$ & 8 \\ \hline
        Optimizer & AdamW \\ \hline
        Number of initial training data $D_\mathrm{init}$ & 100\\ \hline
        Number of iterations $N_\mathrm{iter}$ & 1500 \\ \hline
        Number of epochs & 1000 \\ \hline
        Learning rate & 0.01 \\ \hline
    \end{tabular}

    \vspace{1em} 
        
    \begin{tabular}{|l||c|}
        \hline
        \multicolumn{2}{|c|}{\textbf{Ising machine}} \\ \hline \hline
        Ising solver & Simulated Annealing Sampler\\ \hline
        Initial inverse temperature  & $10^{-5}$ \\ \hline
        Final inverse temperature & $100$ \\ \hline
        Annealing schedule & Linear for the inverse temperatures \\ \hline
        Number of outer loops & 1000 \\ \hline
        Number of inner loops & 10 \\ \hline
        Reads samples per iteration $D_\mathrm{reads}$ & 15 \\ \hline
        Added samples per iteration $D_\mathrm{adds}$ & 3 \\ \hline
        Total added solutions to the dataset & 4600 \\ \hline
    \end{tabular}
\end{table}
\begin{figure*}[t]
\centering
\includegraphics[width=0.8\textwidth]{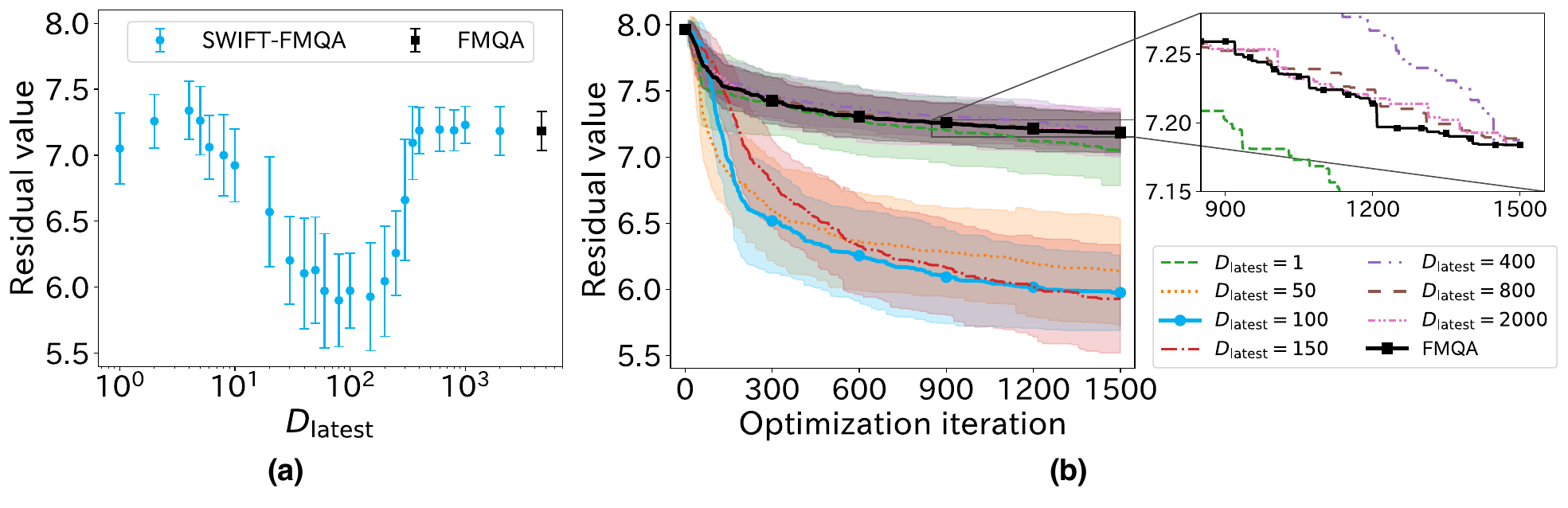}
\caption{Analysis of the residual value with respect to (a) $D_{\mathrm{latest}}$ and (b) optimization iterations.
The markers and lines indicate the mean residual values obtained from $50$ simulations, while the error bars and shaded areas represent their standard deviation.
(a) Dependence of the lowest residual value on $D_{\mathrm{latest}}$, evaluated after $N_{\mathrm{iter}}$ optimization iterations with SWIFT-FMQA completed.
For comparison, the lowest residual value obtained with FMQA is also shown.
(b) Transition of the lowest residual value over the optimization iterations.
For visual clarity, markers are displayed only for the representative cases of SWIFT-FMQA with $D_{\mathrm{latest}}=100$ and FMQA.}
\label{fig:results_Dlatest}
\end{figure*}

As described in Section~\ref{Step 4: constructing the dataset for FM training}, during the optimization process, an Ising solver samples $D_{\mathrm{reads}}$ solutions, and the $D_{\mathrm{adds}}$ solutions with the lowest outputs of the BB function are added to the dataset.
As we defined the dataset as a multiset in Section~\ref{subsection: Procedure of SWIFT-FMQA}, we adopted a strategy that permits duplicates when adding solutions and their corresponding outputs to the dataset.
That is, sampled solutions and their corresponding outputs are added even when they already exist in the previous dataset.
By consistently adding $D_{\mathrm{adds}}$ data points regardless of the duplication, we can guarantee that these new data points always lie within the sliding window.
Furthermore, the resampling of a previously explored solution can be considered as the current FM model still deems that solution to be promising.
Adding such duplicate data points to the dataset might reinforce the learning of information about promising search regions and is also expected to accelerate convergence to the optimal solution.
Throughout the full iteration of the optimization process, the total number of added solutions to the dataset is calculated as $D_{\mathrm{init}} + N_{\mathrm{iter}} \times D_{\mathrm{adds}} = 4600$.
This number also represents the number of data points in the dataset after all optimization iterations by FMQA have been completed.

\section{Results}
\label{section:Results}
In this section, we present results of the simulations with settings described in Section~\ref{section:Simulation settings} to show the advantages of SWIFT-FMQA.
Firstly, with the input dimension of the LABS problem fixed at $N=64$, we show the dependence of the residual value, defined as the difference between the objective function value of the obtained minimum solution and the optimal value ($-9.\dot{8}4615\dot{3}$)~\cite{Packebusch2016Low}, on $D_{\mathrm{latest}}$.
In addition, we confirm that the performance improvement achieved at $N=64$ by a specific choice of $D_{\mathrm{latest}}$ remains consistent across different parameter settings.
To show this property, we carried out simulations by varying the input dimension as $N=16, 49, 64, 81$, and $101$; the hyperparameter as $K=4, 6, 10$, and $16$; and the number of data points in the initial dataset as $D_{\mathrm{init}}=1, 10$, and $100$.

\subsection{\texorpdfstring{$D_{\mathrm{latest}}$}{Dlatest} dependency of the optimization performance}
\label{subsection:D_latest Dependency of the optimization performance}
Fig.~\ref{fig:results_Dlatest}(a) shows the $D_{\mathrm{latest}}$ dependency of the residual values for the problem setting with $N=64$.
Fig.~\ref{fig:results_Dlatest}(b) shows the transition of the lowest residual values over the optimization iterations.
In each simulation, only the initial parameters of FM were varied, while the initial training data and the seed values of the Ising solver were all fixed.
Fig.~\ref{fig:results_Dlatest}(a) shows that, for $10 \leq D_{\mathrm{latest}} \leq 300$, SWIFT-FMQA yields lower residual values than FMQA.
For $D_{\mathrm{latest}}$ outside this range, SWIFT-FMQA produces residual values comparable to those of FMQA.
From Fig.~\ref{fig:results_Dlatest}(b), we observe that for $D_{\mathrm{latest}} = 50$, $100$, and $150$, SWIFT-FMQA achieves the same residual values as FMQA while requiring fewer BB function evaluations.
In particular, SWIFT-FMQA updates the lowest residual value at a faster rate than FMQA during the early stages of the optimization iterations. 
Even in the later stages, it exhibits a steeper improvement in the minimum residual value compared with FMQA.

As shown in Fig.~\ref{fig:results_Dlatest}, the mean residual value is greater than zero for both SWIFT-FMQA and FMQA.
This implies that neither method found the optimal solution for the problem setting with $N=64$ in any of the $50$ simulations.
In contrast, the optimal solutions were obtained for $N=16$, as shown in Appendix~\ref{appendix:Acquisition of the optimal solutions}.

\subsection{Parameters dependency of the optimization performance in \texorpdfstring{$D_{\mathrm{latest}}=100$}{Dlatest=100}}
\label{subsection:FMQA parameters dependency of the optimization performance}
\begin{figure*}[t]
\centering
\includegraphics[width=0.9\textwidth]{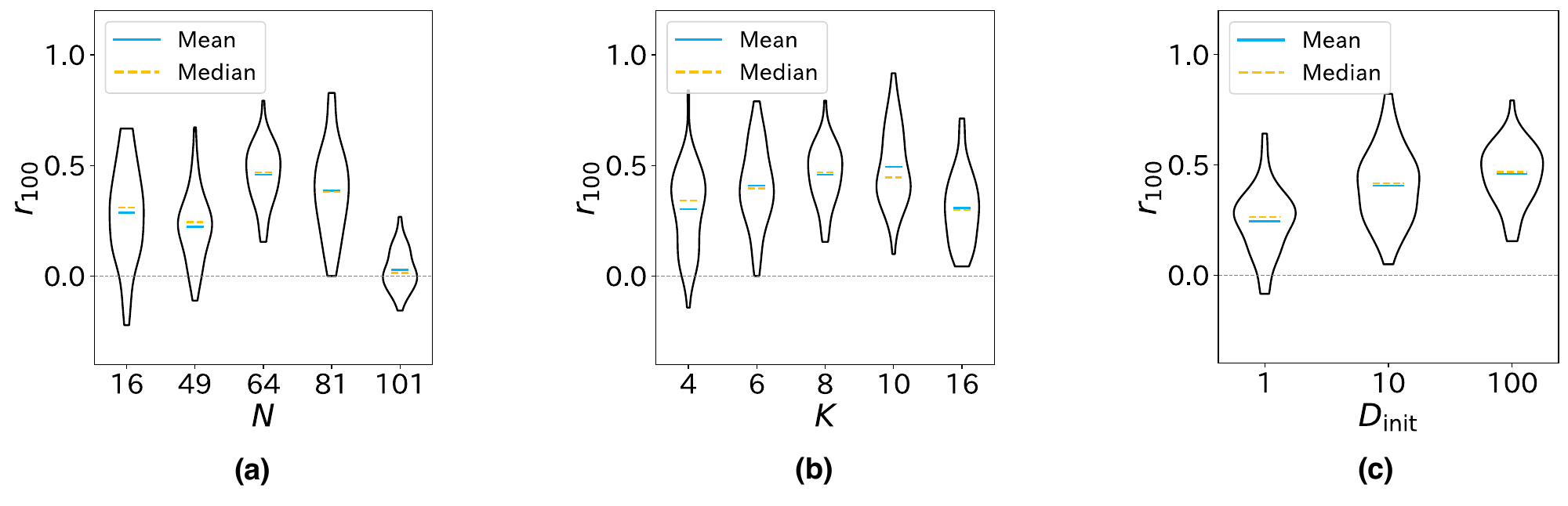}
\caption{Parameter dependence of the improvement rate for $D_{\mathrm{latest}} = 100$. (a) Dependence on the problem size $N$. (b) Dependence on the FM hyperparameter $K$. (c) Dependence on the number of data points used for initial training, $D_{\mathrm{init}}$. For panels (b) and (c), the problem size is fixed at $N = 64$. All parameters other than the one varied along the horizontal axis are set according to Table~\ref{table:Parameter_for_FMQA}.}
\label{fig:results_Dlatest100_comparison}
\end{figure*}
Fig.~\ref{fig:results_Dlatest100_comparison} shows how the improvement rate of the minimum objective function value obtained by SWIFT-FMQA over FMQA depends on the parameters, for $D_{\mathrm{latest}} = 100$ and based on $50$ simulations.
The value $D_{\mathrm{latest}} = 100$ was selected because, as shown in Figs.~\ref{fig:results_Dlatest}(a) and (b), SWIFT-FMQA attains the lowest objective function value in the vicinity of this setting when compared with FMQA.
For $D_{\mathrm{latest}} \neq 100$, the dependence of the lowest residual value on the problem sizes $N$ is discussed in Appendix~\ref{appendix:suitable Dlatest}.
The improvement rate is defined as
\begin{equation}
\label{eq:improvement_rate}
r_{D_{\mathrm{latest}}} \equiv \frac{V_{D_{\mathrm{latest}}}-{V_{\mathrm{FMQA}}}}{{V_{\mathrm{FMQA}}}}.
\end{equation}
Here, $V_{\mathrm{FMQA}}$ denotes the minimum objective function value obtained by FMQA, and $V_{D_{\mathrm{latest}}}$ denotes that obtained by SWIFT-FMQA with a window size of $D_{\mathrm{latest}}$.
Since both $V_{D_{\mathrm{latest}}}$ and $V_{\mathrm{FMQA}}$ are negative, a positive improvement rate indicates that SWIFT-FMQA with window size $D_{\mathrm{latest}}$ attains a lower objective function value than FMQA.
Figs.~\ref{fig:results_Dlatest100_comparison}(a), (b), and (c) show the improvement rate across various problem sizes $N$, FM hyperparameters $K$, and the numbers of data points in the initial training dataset $D_{\mathrm{init}}$, respectively.
All parameter settings, except the parameter being varied in each graph, are followed as described in Table~\ref{table:Parameter_for_FMQA}.

Fig.~\ref{fig:results_Dlatest100_comparison}(a) demonstrates that, for all $N$, SWIFT-FMQA generally identifies solutions whose objective function values are equal to or lower than those obtained by FMQA.
However, the magnitude of the improvement rate exhibits variability.
In particular, in the region where $N$ is large, the improvement rate is found to be low.

Fig.~\ref{fig:results_Dlatest100_comparison}(b) shows that, for the problem size $N = 64$, SWIFT-FMQA samples solutions with lower objective function values than FMQA for all $K$.
For $4 \leq K \leq 10$, the improvement rate increases as $K$ grows, although the rate of change is small.
This result indicates that SWIFT-FMQA does not require precise fine-tuning of the FM hyperparameter $K$ in order to sample solutions with lower objective function values than FMQA.

Fig.~\ref{fig:results_Dlatest100_comparison}(c) illustrates that SWIFT-FMQA obtains solutions with lower objective function values than FMQA across all choices of $D_{\mathrm{init}}$ for the problem size $N=64$.
When $D_{\mathrm{init}}$ is significantly reduced, for example to $D_{\mathrm{init}} = 1$, the improvement rate becomes smaller than that observed at $D_{\mathrm{init}} = 100$, yet the decrease remains modest.
As described in Section~\ref{section:FMQA}, constructing the initial training dataset requires calling the BB function as many times as the number of data points in the dataset, and each BB function evaluation incurs a non-negligible computational cost.
Therefore, the fact that SWIFT-FMQA can produce solutions with lower objective function values than FMQA even for small $D_{\mathrm{init}}$ suggests that it achieves efficient BB optimization with fewer costly evaluations.

\section{Discussion}
\label{section:Discussion}
Firstly, we discuss the optimization mechanism of SWIFT-FMQA and its applicability based on the experimental results.
The results in Section~\ref{section:Results} demonstrated that SWIFT-FMQA with an appropriate $D_{\mathrm{latest}}$ (e.g., $D_{\mathrm{latest}} \simeq 100$) outperforms FMQA. 
We attribute this performance improvement to an appropriate balance between model inertia and generalization capability.
As discussed in Section~\ref{subsection: Motivation: data dilution in FMQA}, FMQA retains all data, which tends to dilute the contribution of newly added data points and consequently results in high model inertia.
Such high inertia may prevent the surrogate model from accurately capturing the shape of low-cost regions in the BB function.
In contrast, retaining too few data points (e.g., $D_{\mathrm{latest}} = 1$) leads to unstable optimization behavior, as observed in Fig.~\ref{fig:results_Dlatest}.
This is likely caused by overfitting, whereby the model loses its predictive generalization ability.
A detailed analysis validating the presence of high model inertia for large $D_{\mathrm{latest}}$ and overfitting for excessively small $D_{\mathrm{latest}}$ is provided in Appendix~\ref{appendix:Verification of high model inertia} and Appendix~\ref{appendix:Analysis of generalization capability of FM}, respectively.
These supplementary analyses suggest that SWIFT-FMQA achieves its effectiveness by maintaining the number of data points in the dataset small enough to remain sensitive to newly sampled solutions with high training agility, yet sufficiently large to avoid overfitting.

Next, we discuss the impact of allowing duplicate data points by treating the dataset as a multiset.
In SWIFT-FMQA, data points added in earlier optimization iterations are progressively discarded according to the sliding-window mechanism.
As described in Section~\ref{subsection: Motivation: data dilution in FMQA}, solutions sampled in the later stages of the optimization iterations are generally more promising for improving optimization performance, even when such solutions have already appeared in previous iterations.
By allowing duplicate data points, SWIFT-FMQA can treat repeatedly sampled solutions as informative signals for model refinement, rather than discarding them solely due to duplication. 
From a theoretical perspective, excessive duplication may cause the data points in the dataset to become overly homogeneous, which can in turn lead to vanishing gradients of the loss function of FM.
However, as shown in Appendix~\ref{appendix:Verification of high model inertia}, the gradients of the loss function in SWIFT-FMQA remain sufficiently large throughout the optimization iterations under our experimental settings.
This indicates that gradient vanishing does not occur and that the model updates remain active.
On this basis, duplicate data points are allowed in this work.

Finally, we discuss the limitations of SWIFT-FMQA concerning the benchmark problem and highlight its practical implications for real-world, resource-constrained optimization tasks.
As noted in the results for $N=64$ (Fig.~\ref{fig:results_Dlatest}), although SWIFT-FMQA outperformed FMQA, it did not reach the global optimum obtained by exhaustive search.
This limitation is largely attributable to the structural mismatch discussed in Section~\ref{subsection: Black-box function}; the quadratic FM model cannot perfectly capture the landscape of the quartic LABS problem.
Consequently, there is an inherent upper bound on the accuracy achievable by any FM-based approach for this specific benchmark.

However, from a practical perspective, our result highlights the strength of SWIFT-FMQA.
In real-world BB optimization scenarios, such as material discovery or structure design, the true global optimum is unknown, and the number of allowed BB function evaluations is strictly limited due to high costs.
The primary goal in such contexts is not necessarily to guarantee the global optimum, but to find better solutions than conventional BB optimization methods within a limited budget.
Fig.~\ref{fig:results_Dlatest}(b) demonstrates that SWIFT-FMQA consistently achieves lower objective function values with fewer BB function evaluations compared to FMQA.
Furthermore, Fig.~\ref{fig:results_Dlatest100_comparison}(a) confirms that SWIFT-FMQA generally obtains solutions equal to or better than FMQA across a range of problem sizes $N$, particularly showing a clear advantage for $N \le 81$.
Consequently, SWIFT-FMQA offers practical utility as a cost-effective solver for complex problems where resource constraints are a critical factor, at least up to moderate problem sizes.
Improving the scalability of SWIFT-FMQA remains an important direction for future work.

\begin{table*}[t]
    \centering
    \caption{Number of times the optimal solution was obtained out of $50$ simulations for $N=16$.}
    \label{table:acquisition of the optimal solutions}
    \renewcommand{\arraystretch}{1.15}
    \begin{tabularx}{0.8\textwidth}{|c||*{13}{Y|}}
        \hline
        & \multicolumn{13}{c|}{\textbf{SWIFT-FMQA}} \\
        \hline
        $D_{\mathrm{latest}}$ & 1 & 2 & 4 & 5 & 6 & 8 & 10 & 20 & 30 & 40 & 50 & 60 & 80\\
        \hline
        Count & 5 & 3 & 3 & 3 & 4 & 5 & 6 & 1 & 6 & 2 & 5 & 5 & 12\\
        \hline
        \hline
        & \multicolumn{11}{c|}{\textbf{SWIFT-FMQA}} & \multicolumn{2}{c|}{\multirow{2}{*}{\textbf{FMQA}}}\\
        \cline{1-12}
        $D_{\mathrm{latest}}$ & 100 & 150 & 200 & 250 & 300 & 350 & 400 & 600 & 800 & 1000 & 2000 & \multicolumn{2}{c|}{} \\
        \hline
        Count & 27 & 44 & 45 & 46 & 42 & 45 & 40 & 38 & 25 & 26 & 16 & \multicolumn{2}{c|}{1} \\
        \hline
    \end{tabularx}
\end{table*}
\section{Conclusion}
\label{section:Conclusion}
The main contributions of this study are summarized as follows:

\begin{itemize}
    \item We proposed a novel BB optimization method named SWIFT-FMQA that employs a sliding-window strategy to sequentially construct the training dataset. By retaining only the most recently added data points up to a specified limit ($D_{\mathrm{latest}}$), this method prevents data dilution and mitigates high model inertia.
    
    \item We demonstrated the superior optimization performance of SWIFT-FMQA through numerical experiments on the LABS problem. The results showed that SWIFT-FMQA consistently finds lower-cost solutions with fewer BB function evaluations compared to FMQA.
    
    \item We revealed that SWIFT-FMQA exhibits robustness against hyperparameter settings. Specifically, it achieves high optimization performance without requiring sensitive fine-tuning of the FM hyperparameter $K$ or the number of data points in the initial dataset $D_{\mathrm{init}}$.
\end{itemize}

\appendices
\section{\break{Acquisition of the optimal solutions}}
\label{appendix:Acquisition of the optimal solutions}
In this section, we show the frequency of obtaining optimal solutions through BB optimization by FMQA and SWIFT-FMQA.
As described in Section~\ref{section:Results}, we found that SWIFT-FMQA can search for solutions with lower residual values with respect to the optimal solution compared to FMQA.
However, as shown in Figs. \ref{fig:results_Dlatest}(a) and (b), for the problem setting with $N=64$, only solutions with a residual value of approximately $5.8$ or higher were obtained for all $D_{\mathrm{latest}}$.
As mentioned in Section~\ref{subsection: Black-box function}, for the LABS problem, the number of candidate solutions increases with the number of input variables $N$, making it difficult to find the optimal solution.
We confirm whether the optimal solution can be sampled by reducing $N$, which simplifies the BB function setting.

For the problem settings with $N=16, 49$, and $64$ selected in Fig.~\ref{fig:results_Dlatest100_comparison}(a), we examined the number of times the optimal solutions were obtained out of $50$ simulations.
As a result, the optimal solutions were obtained only for $N=16$.
Table~\ref{table:acquisition of the optimal solutions} shows the dependence on $D_{\mathrm{latest}}$ of the number of times the optimal solution was obtained for $N=16$.
Note that for $N=16$, there are four optimal solutions~\cite{Packebusch2016Low} and their objective function values are $-5.\dot{3}$.
We counted the number of times the optimal solution was obtained, regardless of which particular optimal solution was obtained.
As shown in Table~\ref{table:acquisition of the optimal solutions}, it was found that for all $D_{\mathrm{latest}}$, SWIFT-FMQA matched or exceeded FMQA in the number of times the optimal solution was obtained.
Furthermore, for $150 \leq D_{\mathrm{latest}} \leq 400$, we found that the optimal solutions were obtained $40$ or more times out of $50$ simulations.
These results indicate that SWIFT-FMQA is an effective method for obtaining the optimal solutions to BB optimization problems, which are difficult to find using FMQA.

\section{\break{ \texorpdfstring{$D_{\mathrm{latest}}$}{Dlatest} dependency on the problem size \texorpdfstring{$N$}{N}}}
\label{appendix:suitable Dlatest}
\begin{figure*}[t]
\centering
\includegraphics[width=0.7\textwidth]{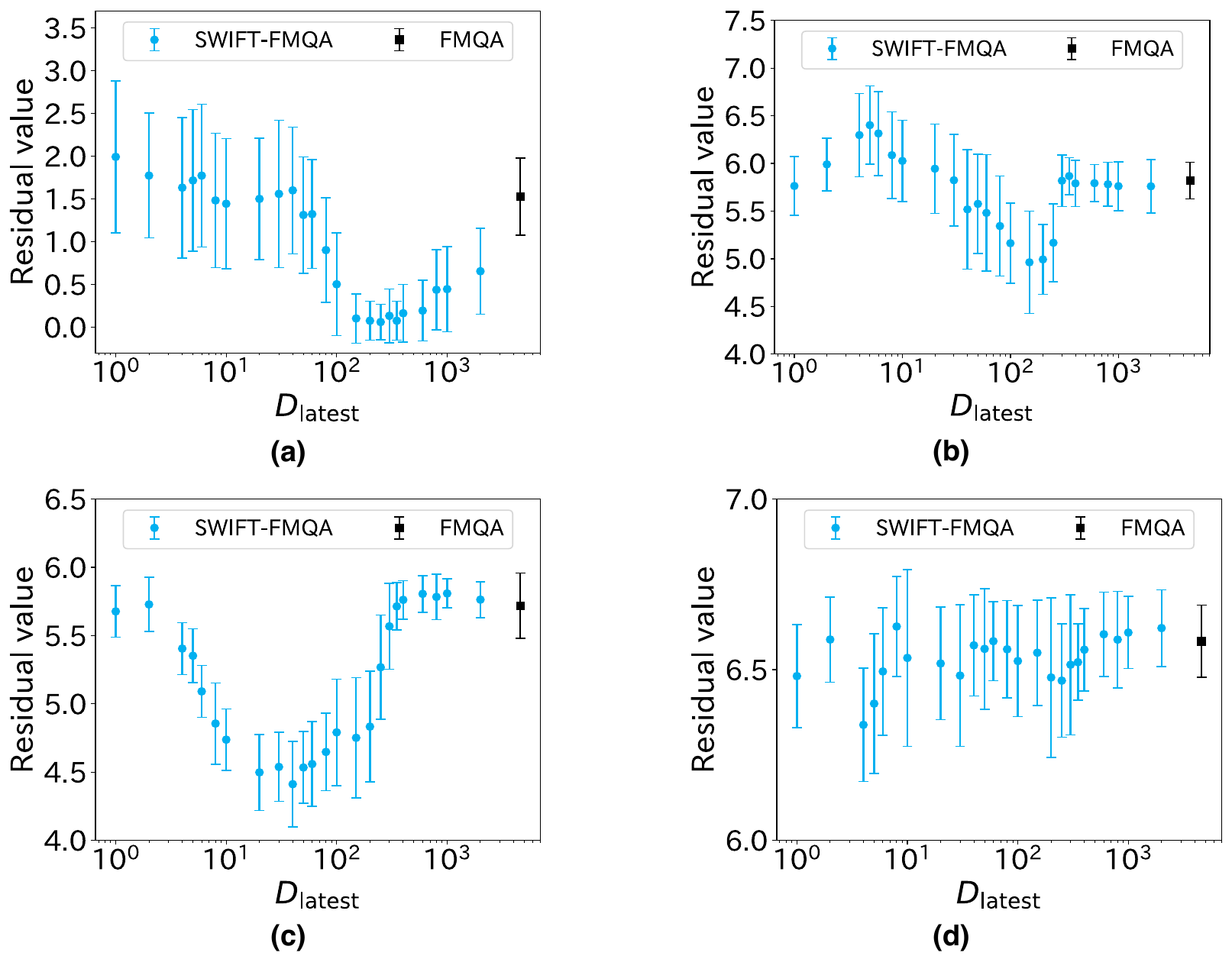}
\caption{$D_{\mathrm{latest}}$ dependency of the obtained lowest residual value until the $N_{\mathrm{iter}}$ optimization iterations have been completed. The error bars indicate the standard deviation of the residual values obtained from $50$ simulations. (a) $N=16$. (b) $N=49$. (c) $N=81$. (d) $N=101$.}
\label{fig:various_N_results_Dlatest}
\end{figure*}
In this section, we discuss the $D_{\mathrm{latest}}$ dependency for various problem sizes $N$ to obtain solutions with lower residual values than FMQA.
Fig.~\ref{fig:various_N_results_Dlatest} shows the dependence of the residual values on $D_{\mathrm{latest}}$ for $N=16, 49, 81$, and $101$, obtained from $50$ simulations.
Here, the residual value is defined as the difference between the objective function value of the obtained minimum solution and the best-known value reported in the literature~\cite{Packebusch2016Low}.
We found that for every $N$, there are suitable $D_{\mathrm{latest}}$ settings that obtain lower residual values than FMQA.
Furthermore, it was found that the best setting for $D_{\mathrm{latest}}$ to obtain the lowest residual value among all $D_{\mathrm{latest}}$ tends to be different for $N$.
The optimal $D_{\mathrm{latest}}$ increases as $N$ decreases, and vice versa.

These results consistently indicate that the optimal $D_{\mathrm{latest}}$ lies between the two extremes: the case where the number of data points in the dataset is excessively small (e.g., $D_{\mathrm{latest}}=1$) and the case where it is excessively large (i.e., FMQA).
While our experiments demonstrate this relationship, deriving a generalized formula or theoretical framework to determine the precise optimal $D_{\mathrm{latest}}$ based on problem characteristics is beyond the scope of this paper.
A detailed analysis of the energy landscape to elucidate the quantitative relationship between $N$ and the optimal $D_{\mathrm{latest}}$ remains an important topic for future work.

\section{\break{Verification of high model inertia}}
\label{appendix:Verification of high model inertia}
In this section, we verify that high model inertia arises from a large number of data points in the dataset, as discussed in Section~\ref{subsection: Motivation: data dilution in FMQA}.
Model inertia is inversely related to the model’s sensitivity to newly added data, such that lower sensitivity corresponds to higher inertia.
Accordingly, we introduce a metric termed \textit{training agility} to quantify the sensitivity of the FM model in~\eqref{eq:FM-general}.

Training agility is defined as the norm of the gradient of the loss function with respect to the model parameters, evaluated immediately after a dataset update.
Let $\bm{\theta}^{(t)}$ be the model parameters of FM trained with the dataset $\mathcal{D}^{(t-1)}$ at iteration index $t$.
After the training of FM, an Ising solver samples new solutions, and the dataset is updated to $\mathcal{D}^{(t)}$ by adding newly sampled solutions (and removing old ones in the case of SWIFT-FMQA).
FM is then retrained on the updated dataset $\mathcal{D}^{(t)}$ at iteration index $t+1$.
Let $L(\bm{\theta}; \mathcal{D}^{(t)})$ denote the corresponding loss function.
To quantify the impact of the dataset update, we evaluate the gradient of $L(\bm{\theta}; \mathcal{D}^{(t)})$ with respect to $\bm{\theta}$ at the fixed parameter value $\bm{\theta} = \bm{\theta}^{(t)}$.
The gradient is expressed as:
\begin{equation}
\label{eq:loss_func_gradient_vector}
\nabla_{\bm{\theta}} L = \frac{2}{D} \sum_{d=1}^{D} \left( f_{\mathrm{FM}}(\bm{x}^{(d)}; \bm{\theta}) - f(\bm{x}^{(d)}) \right) \frac{\partial f_{\mathrm{FM}}(\bm{x}^{(d)}; \bm{\theta})}{\partial \bm{\theta}}.
\end{equation}
The training agility is defined as the $L^2$ norm of the gradient of the loss function:
\begin{equation}
\label{eq:loss_func_L2}
\| \nabla_{\bm{\theta}} L \|_2 = \sqrt{\left(\frac{\partial L}{\partial \omega_0}\right)^2 + \sum_{i=1}^N \left(\frac{\partial L}{\partial \omega_i}\right)^2 + \sum_{i=1}^N \|\nabla_{\bm{v}_i} L \|_2^2}.
\end{equation}
Based on the model equation of FM given in \eqref{eq:FM-general}, the partial derivatives with respect to each model parameter required for \eqref{eq:loss_func_gradient_vector} can be derived explicitly as follows:
\begin{subequations}
\label{eq:FM_derivatives}
\begin{align}
\frac{\partial f_{\mathrm{FM}}}{\partial \omega_0} &= 1, \\
\frac{\partial f_{\mathrm{FM}}}{\partial \omega_i} &= x_i, \\
\frac{\partial f_{\mathrm{FM}}}{\partial v_{i,l}} &= x_i \left(\sum_{j=1}^N v_{j,l} x_j\right)-v_{i, l}x_i^2.
\end{align}
\end{subequations}
Here, $v_{i,l}$ denotes the $l$-th component of the vector $\bm{v}_i$.
A large training agility indicates that the model parameters $\bm{\theta}^{(t)}$ remain highly sensitive to newly added data, leading to substantial updates of the model.
Conversely, a small training agility implies that the new data has little influence on the model, thereby indicating high model inertia.

\begin{figure}[t]
\centering
\includegraphics[width=0.3\textwidth]{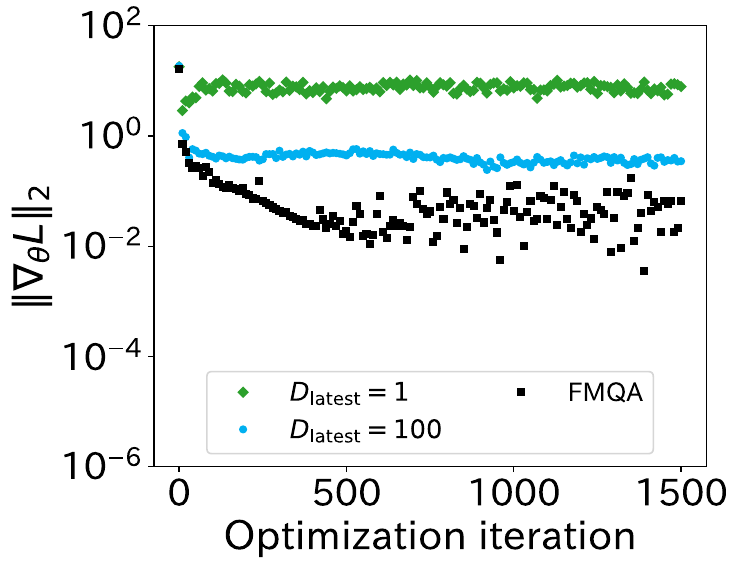}
\caption{Mean training agility over 50 simulations versus optimization iterations.}
\label{fig:training_gradient_norm}
\end{figure}

Fig.~\ref{fig:training_gradient_norm} shows the $D_{\mathrm{latest}}$ dependency of the transition of the training agility for the problem setting with $N=64$ obtained from $50$ simulations.
For FMQA, the training agility decreases by nearly two orders of magnitude from the initial stage to the $1500$th iteration.
This observation provides empirical evidence of high model inertia resulting from data dilution discussed in Section~\ref{subsection: Motivation: data dilution in FMQA}.
In contrast, SWIFT-FMQA with $D_{\mathrm{latest}} = 100$ consistently maintains a moderate level of training agility throughout the optimization process, compared with FMQA.
This demonstrates that employing a sliding-window strategy enables the model to retain sensitivity to newly added solutions.
From this observation, we conclude that the strategy of replacing duplicated solutions with randomly generated ones, as mentioned in Section~\ref{subsection: Procedure of SWIFT-FMQA}, is not required for SWIFT-FMQA under the experimental settings considered in this study.
In the extreme case of SWIFT-FMQA with $D_{\mathrm{latest}} = 1$, the training agility reaches the highest values, often exceeding $10$, which is consistent with expectations from the perspective of data dilution.

\section{\break{Analysis of generalization capability of FM}}
\label{appendix:Analysis of generalization capability of FM}
In this section, we investigate the relationship between the number of data points in the dataset and the generalization capability of FM for $N=64$.

\begin{figure}[t]
\centering
\includegraphics[width=0.3\textwidth]{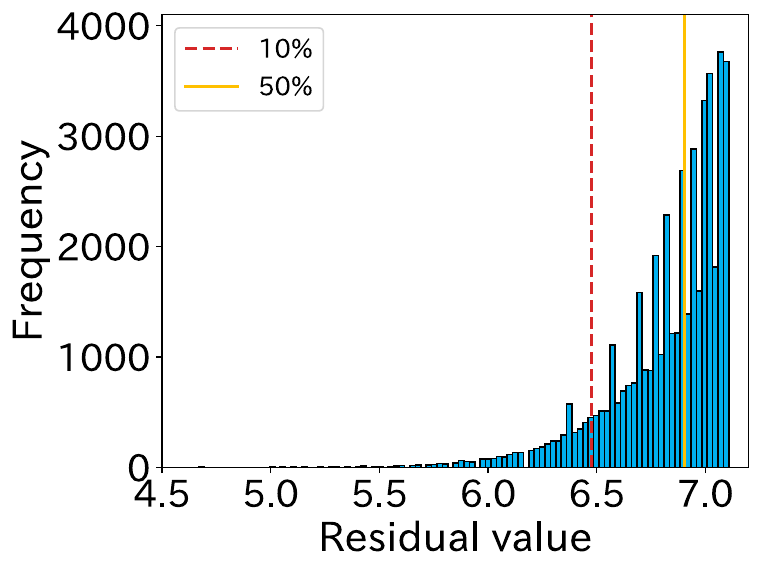}
\caption{Data pool including training data and test data for analysis of generalization capability of FM.}
\label{fig:training_data_hist}
\end{figure}

Firstly, we prepared a large set of evaluated solutions consisting of $5.75$ million input-output pairs of binary variables and their corresponding objective function values.
These pairs were generated through $50$ independent optimization simulations using both FMQA and SWIFT-FMQA across $25$ different $D_{\mathrm{latest}}$ values shown in Fig.~\ref{fig:results_Dlatest}(a), following the parameter settings in Table~\ref{table:Parameter_for_FMQA}.
From this set, we constructed a data pool by selecting $46000$ unique solutions with the lowest objective function values.
The distribution of residual values within the constructed data pool is presented in Fig.~\ref{fig:training_data_hist}.
The data is most concentrated around a residual value of $7.0$, while the number of samples decreases significantly as the residual value becomes smaller.

We evaluated the training and test errors of FM using the constructed data pool.
The evaluation was conducted across $10$ independent trials to ensure statistical reliability.
In each trial, the number of data points in the training dataset $D$ was varied (e.g., $D=1$, $100$, and $4600$), where the data points were randomly sampled from the pool.
Note that $D=4600$ corresponds to the number of data points in the dataset after all optimization iterations by FMQA have been completed.
The number of data points in the test dataset was fixed at $1150$, so as to maintain an $8:2$ split relative to the maximum of $4600$ data points used in the training dataset.
The test data were also randomly sampled from the pool, ensuring no overlap with the training data.
To strictly verify the prediction accuracy of the FM across different regions of the energy landscape, particularly near low-cost solutions, we prepared three types of test datasets for test: sampling from the entire data pool (Top $100\%$), the top $50\%$ of solutions with the lowest objective function values, and the top $10\%$ of them.
The configuration of FM was consistent with Table~\ref{table:Parameter_for_FMQA}, except for the number of epochs.
The number of epochs was set to $2000$ to ensure sufficient convergence of the model parameters.
We used the MSE, defined in \eqref{eq:learning_loss}, as the metric for both training errors and test errors.
\begin{figure*}[t]
\centering
\includegraphics[width=1.0\textwidth]{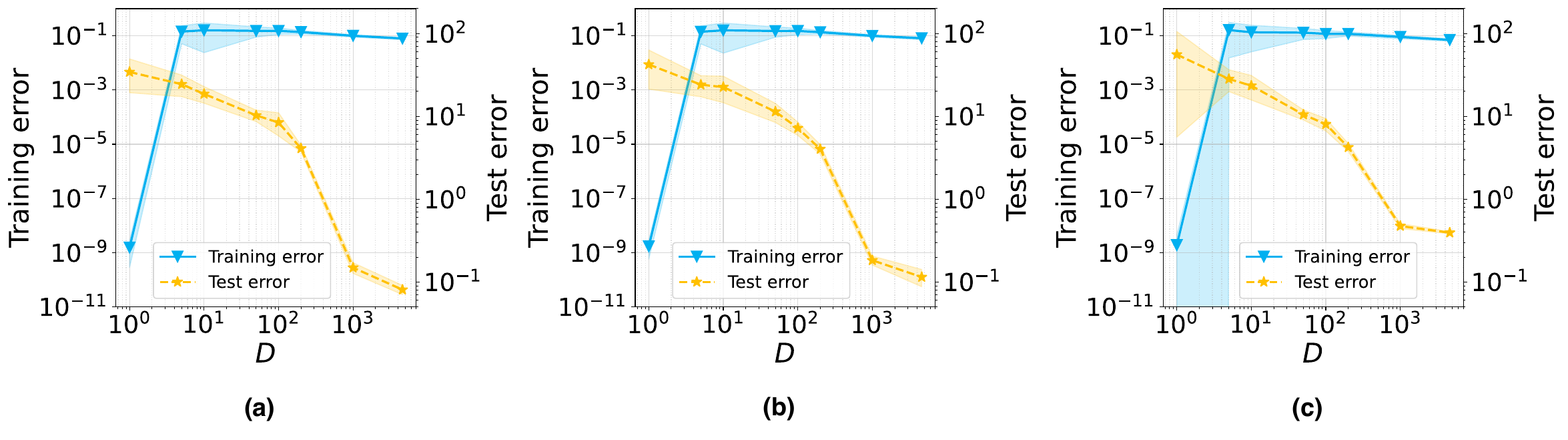}
\caption{Training and test errors as functions of the number of data points in the dataset. Markers represent the mean values, and shaded areas represent the standard deviation obtained from $10$ simulations. The solid blue lines represent the training error, while the dashed yellow lines represent the test errors. The test data are randomly sampled from the data pool, ensuring no overlap with the training data. (a) Test data were sampled from the entire data pool. (b) Test data were sampled from top $50\%$ solutions with the lowest objective function values. (c) Test data were sampled from top $10\%$ solutions with the lowest objective function values. Note that the rightmost point in each plot corresponds to the number of data points in the dataset after all optimization iterations by FMQA have been completed ($D=4600$).}
\label{fig:train_test_error}
\end{figure*}

Fig.~\ref{fig:train_test_error} clearly illustrates the critical trade-off between the number of data points in the dataset and the model’s generalization capability.
In the regime where $D$ is extremely small (e.g., $D = 1$), the training error becomes negligibly small, whereas the test error remains high across all test datasets.
This pronounced generalization gap is indicative of overfitting: the model perfectly memorizes the limited available data points but fails to capture the landscape of the BB function.
This observation provides a quantitative explanation for the poor optimization performance of SWIFT-FMQA with $D_{\mathrm{latest}} = 1$ observed in Fig.~\ref{fig:results_Dlatest}.
Although the model exhibits high training agility, as shown in Fig.~\ref{fig:training_gradient_norm}, its search direction becomes unreliable due to insufficient predictive generalization.

As the number of data points in the dataset increases, the test error decreases, indicating a substantial improvement in generalization capability.
In particular, when $D$ becomes extremely large (e.g., $D = 4600$), the test error attains its lowest value among all test datasets, confirming that incorporating all available data yields the most predictive model.
However, despite this superior generalization performance, FMQA still exhibits optimization stagnation, as shown in Fig.~\ref{fig:results_Dlatest}.
This comparison reinforces our conclusion in Section~\ref{section:Discussion} that the stagnation observed in FMQA is not attributable to insufficient prediction accuracy, but rather to excessive model inertia.
Consequently, the success of SWIFT-FMQA with $D_{\mathrm{latest}} \simeq 100$ can be attributed to an appropriate balance between model inertia and generalization capability.

\section*{Acknowledgment}
Human Biology-Microbiome-Quantum Research Center (Bio2Q) is supported by World Premier International Research Center Initiative (WPI), MEXT, Japan.
We are grateful to Masashi Yamashita and Tokiya Fukuda for their collaboration in the early stages of this work.

\begin{IEEEbiography}[{\includegraphics[width=1in,height=1.25in,clip,keepaspectratio]{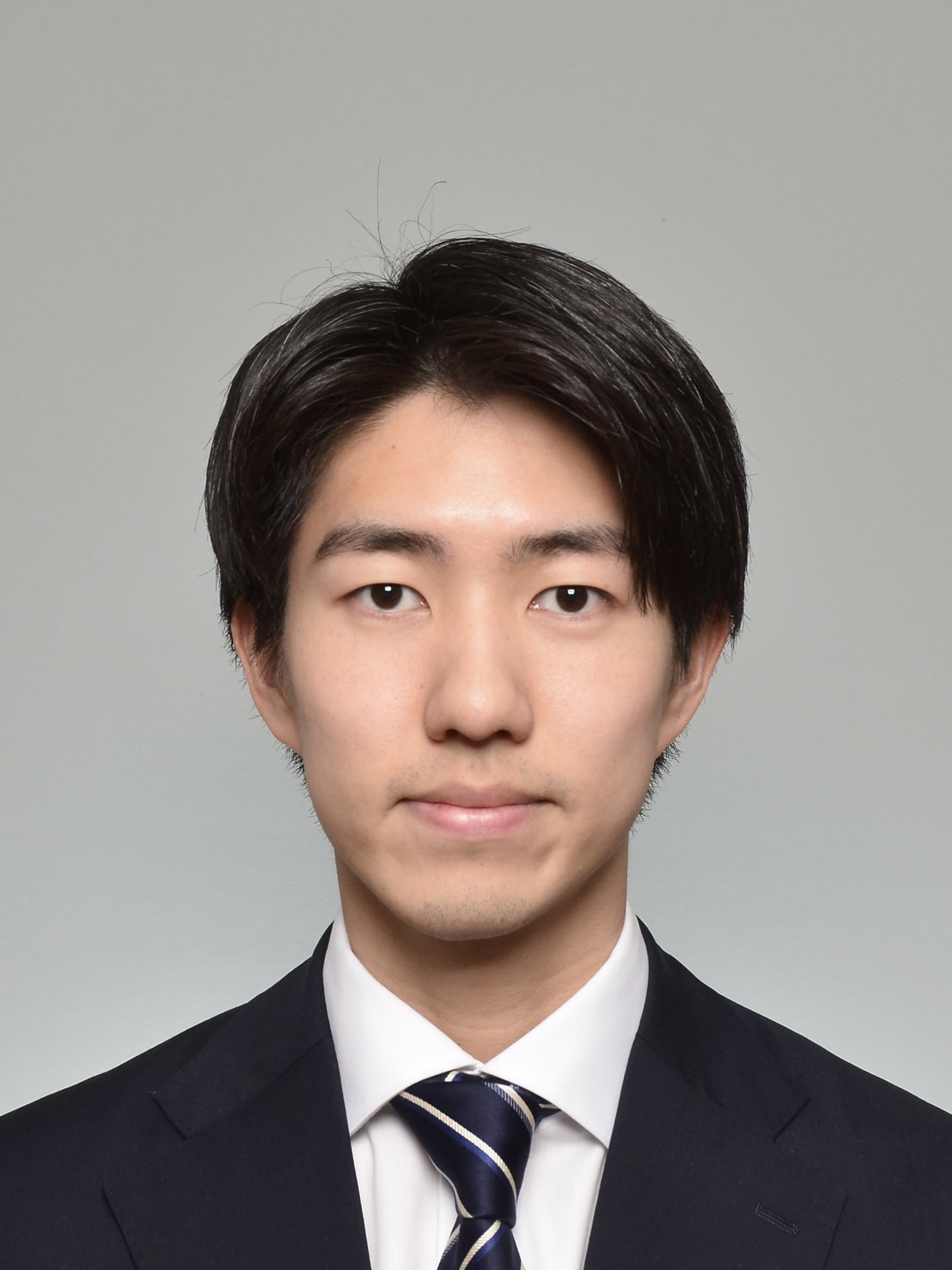}}]{Mayumi Nakano} 
received the B.~Eng. degree in applied physics and physico-informatics from Keio University, Kanagawa, Japan, in 2024, where he is currently pursuing the M.~Eng. degree in fundamental science and technology.
His research interests include black-box optimization, mathematical optimization, machine learning,  quantum annealing, and Ising machines.
\end{IEEEbiography}
\begin{IEEEbiography}[{\includegraphics[width=1in,height=1.25in,clip,keepaspectratio]{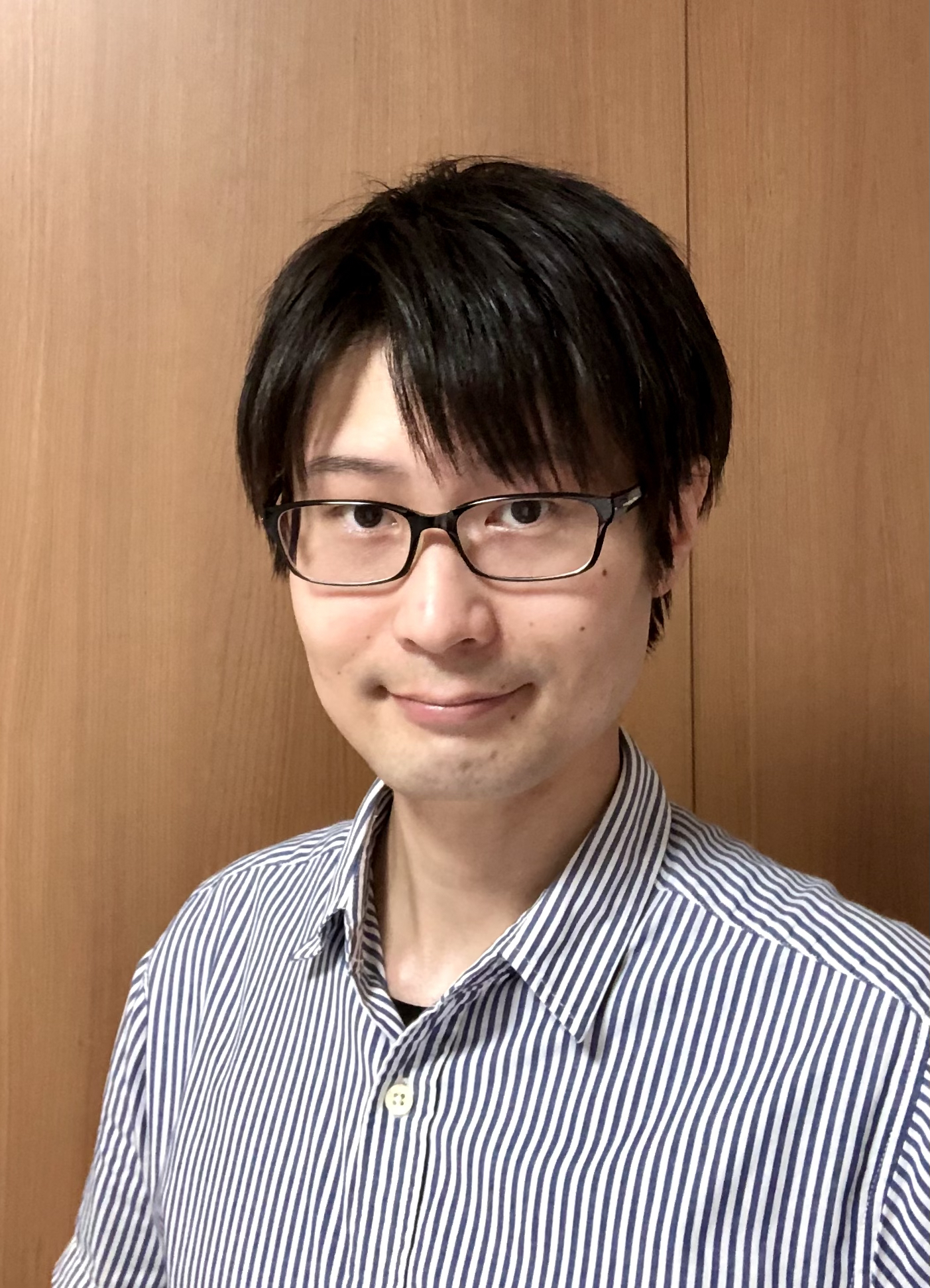}}]{Yuya Seki}
received the B.~Sc., M.~Sc. and Dr.~Sc. degrees from the Tokyo Institute of Technology, in 2011, 2013, and 2016, respectively.
He is currently a Project Lecturer with the Graduate School of Science and Technology, Keio University.
His research interests include quantum computing, machine learning, statistical physics, quantum annealing, and Ising machines.
He is a member of the Physical Society of Japan (JPS).
\end{IEEEbiography}
\begin{IEEEbiography}[{\includegraphics[width=1in,height=1.25in,clip,keepaspectratio]{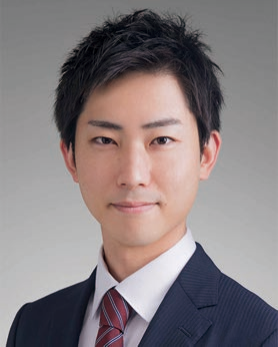}}]{Shuta Kikuchi} 
received the B.~Eng. and M.~Eng. degrees from Waseda University, Tokyo, Japan, in 2017 and 2019, and Dr.~Eng. degrees from Keio University, Kanagawa, Japan in 2024.
He is currently a Project Assistant Professor with the Graduate School of Science and Technology, Keio University.
His research interests include Ising machine, statistical mechanics, and quantum annealing.
He is a member of the Physical Society of Japan (JPS).
\end{IEEEbiography}
\begin{IEEEbiography}[{\includegraphics[width=1in,height=1.5in,clip,keepaspectratio]{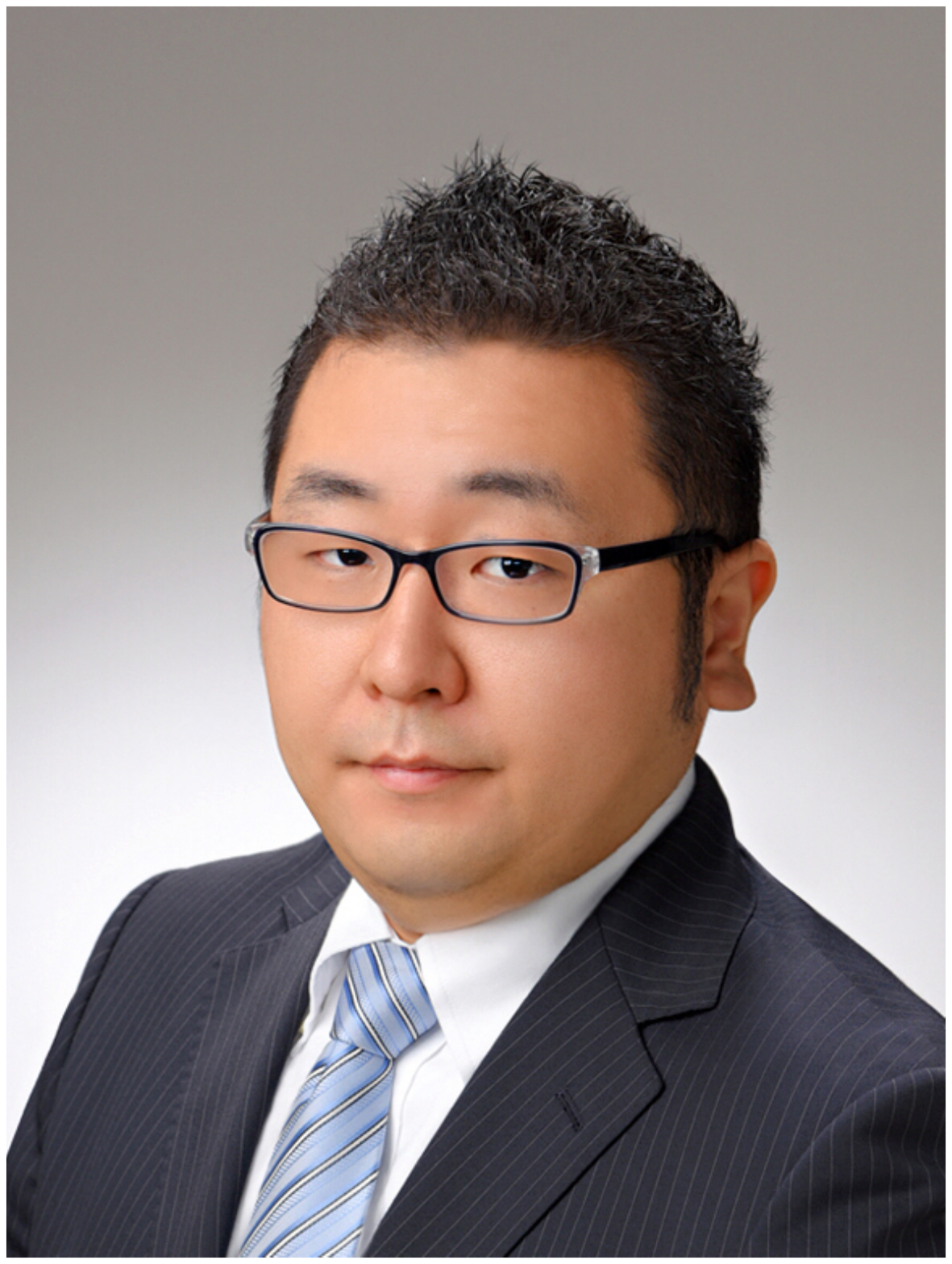}}]{Shu Tanaka} (Member, IEEE)
received his B.~Sci. degree from the Tokyo Institute of Technology, Tokyo, Japan, in 2003, and his M.~Sci. and Ph.~D. degrees from the University of Tokyo, Tokyo, Japan, in 2005 and 2008, respectively.
He is currently a Professor in the Department of Applied Physics and Physico-Informatics, Keio University, a chair of the Keio University Sustainable Quantum Artificial Intelligence Center (KSQAIC), Keio University, and a Core Director at the Human Biology-Microbiome-Quantum Research Center (Bio2Q), Keio University.
His research interests include quantum annealing, Ising machines, quantum computing, statistical mechanics, and materials science.
He is a member of the Physical Society of Japan (JPS), and the Information Processing Society of Japan (IPSJ).
\end{IEEEbiography}
\EOD
\end{document}